\documentclass{article} 
\usepackage{iclr2021_conference,times}

\usepackage{amsmath,amsfonts,bm}

\def\eqref#1{equation~\ref{#1}}

\def\1{\bm{1}}

\DeclareMathAlphabet{\mathsfit}{\encodingdefault}{\sfdefault}{m}{sl}
\SetMathAlphabet{\mathsfit}{bold}{\encodingdefault}{\sfdefault}{bx}{n}

\usepackage{hyperref}
\usepackage{url}

\title{Privacy-preserving object detection}

\usepackage[utf8]{inputenc}
\usepackage[T1]{fontenc}
\usepackage{amsfonts}
\usepackage{amsmath}
\usepackage{accents}
\usepackage{amssymb}
\usepackage{bbm}
\usepackage{bm}
\usepackage{ulem}
\usepackage{booktabs}
\usepackage{chngcntr}
\usepackage{epsfig}
\usepackage{fontawesome}
\usepackage{graphicx}
\usepackage{mathtools}
\usepackage{nicefrac}
\usepackage{pifont}
\usepackage{subcaption}
\usepackage{tabularx}
\usepackage{todonotes}
\usepackage{url}
\usepackage{wrapfig}

\usepackage{hyperref}
\usepackage[capitalise,nameinlink]{cleveref}
\usepackage{array, booktabs, tabularx} 
\usepackage{setspace} 
\usepackage{multirow} 
\usepackage{rotating}
\usepackage{tabularx}
\usepackage{colortbl}
\usepackage{arydshln}

\usepackage{listings}
\usepackage{tikzsymbols}
\usepackage{xcolor}

\newcommand{\midsepremove}{\aboverulesep = 0mm \belowrulesep = 0mm}

\makeatletter
\renewcommand{\paragraph}{%
\@startsection{paragraph}{4}%
{\z@}{0.25em}{-1em}%
{\normalfont\normalsize\bfseries}%
}
\makeatother
\author{
  Peiyang He$^{1}$\thanks{Joint first authors. $^\dagger:\!$ To whom correspondence should be addressed: yuki@robots.ox.ac.uk}, \quad Charlie Griffin,$^{1,\ast}$  \quad  Krzysztof Kacprzyk,$^{1,\ast}$ \quad Artjom Joosen,$^{1}$ \\ 
  \\
  \textbf{Michael Collyer},$^{1}$  \quad \textbf{Aleksandar Shtedritski,}$^{1}$ \quad \textbf{Yuki M. Asano}$^{1}$\footnote{} \vspace{1mm} \\
   $^1$ Oxford Artificial Intelligence Society, University of Oxford \\ 
}
\iclrfinalcopy

\begin{document}
\maketitle
\begin{abstract}
Privacy considerations and bias in datasets are quickly becoming high-priority issues that the computer vision community needs to face. So far, little attention has been given to practical solutions that do not involve collection of new datasets. In this work, we show that for object detection on COCO, both anonymizing the dataset by blurring faces, as well as swapping faces in a balanced manner along the gender and skin tone dimension,  can retain object detection performances while preserving privacy and partially balancing bias.
\end{abstract}

\raggedbottom

\section{Introduction}
Deep learning-based methods have enjoyed tremendous gains over the past years. 
While this has also been due to better architectures, a large part of this success is due to the ever-increasing size of datasets.

However, there are two major problems with the current datasets that the research community is increasingly becoming aware of and that could limit the progress in this domain if not addressed.

First, datasets are being revealed to be heavily biased. Works such as \citep{buolamwini_gebru2018,yang2020towards} find significant underrepresentation of women and those with darker skin in common datasets. 

Biases in datasets, in turn, manifest as bias within models. 
For example, an image labelling algorithm from Google was found to label an image of two black people "Gorilla" \citep{simonite2018google} and facial analysis models have been found to have significantly lower accuracy on darker female faces than on lighter male faces~\citep{buolamwini_gebru2018}.

A second problem is the lack of consent for using these images to train AI models: 
As noted in~\citep{prabhu2020large} while these datasets are collected under the Creative Commons licence, this license does not yield or say anything about their use in training AI models. 
This can potentially violate the licenses (as reproduction requires the ascription of the creator) since trained models can be probed to reveal whole training samples, such as addresses and bank accounts in GPT-2~\citep{carlini2020extracting} or potentially even images~\citep{orekondy18gradient}. 
Therefore, there is a clear need for the removal of personally identifiable information within images. 
This would not only protect the individuals within the images (e.g. reduced risk of identify theft) but also ensure higher adherence to the General Data Protection Regulation (GDPR).

One common variable in both problems is the usage of raw, unedited data which tends to be either privacy infringing or biased.
However, since collecting an unbiased dataset with consenting individuals would be very costly and potentially even unfeasible~\citep{yang2020towards}, in this paper we investigate approaches to mitigating these problems by modifying the current training data. 
Specifically, our contributions are as follows:
\begin{enumerate}
    \item We investigate privacy-preserving measures on object detection performance.
    \item We develop a novel method for balancing gender and race distribution in the training data while simultaneously removing personally-identifying information.
    \item We test the fine-tuned models with two measures of model bias. 
\end{enumerate}

\section{Related work}
\paragraph{Measuring Bias.}
Experimentally, it has been shown that many models are biased: For example~\cite{buolamwini_gebru2018} show that face recognition tools perform worse for women of colour than white men, and in~\citep{simonite2018google, kayserbril2020googlevision} publicly available models have been shown to be biased towards minorities.

Furthermore, \citet{steed2021biases} find that unsupervised models trained on ImageNet contain racial, gender, and intersectional biases.  We use their test to probe bias in our models.

To mitigate biases, \cite{zhang2016discrimination} develop a method to remove discriminatory effects of collected datasets prior to its statistical analysis. 

Even with a balanced dataset, a model can amplify implicit gender biases, as shown by \cite{wang2019balanced}, who propose to use a generative-adversarial approach to remove this information by painting over that part of the image. 
In this paper, we focus on how input data can be efficiently transformed to make datasets more balanced and potentially remove the association biases of stereotypes.
\paragraph{Preserving Privacy.}
Privacy is important within computer vision given the potential contradiction between using image recognition algorithms while at the same time limiting elements which expose identifiable and sensitive information. Preserving privacy in computer vision models has already been addressed to an extent using various methods, such as head inpainting~\citep{Sun2018inpainting}, reducing resolution quality~\citep{Ryoo2016lowres}, and many more~\citep{ren2018privacy, Wu2018preservingprivacy}. Notably, \citep{Orekondy2018privacy} showed how to apply targeted obfuscation to areas of private information as to protect privacy yet also preserve the utility of the image. However, so far little work has evaluated performances of preserving privacy on modern object detection algorithms.
\paragraph{Face Swapping.}
There has also been work in the area of transferring a face onto a target in a similar pose. \cite{Zhong2016} paste celebrity faces onto images to generate a synthetic dataset of novel (celebrity face, action) pair images.
\paragraph{Mitigating bias.}
In \citep{wang2019balanced}, the authors find that performance on multi-label prediction on COCO decreases only slightly when augmenting the model architecture with an adversarial loss to blur parts of the image that cause leakage of protected characteristics such as race and gender.
\begin{figure*}[t]
	\centering
	\includegraphics[width=\textwidth]{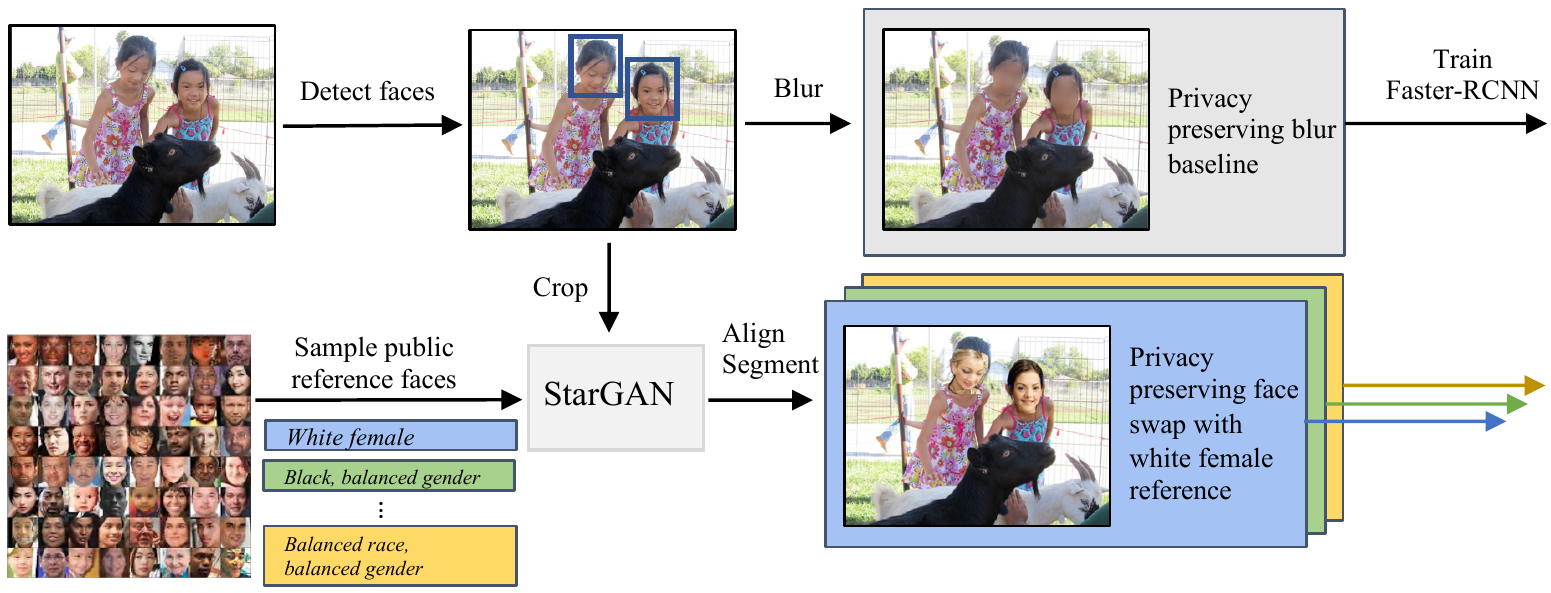}
	\caption{\textbf{Schematic overview of privacy-preserving approaches studied.} We explore various methods such as blurring faces and aligning and pasting faces fom a face dataset with known characteristics. Example image is taken from  MS-COCO (ID=8690).}
	\label{fig:data_gen}
\end{figure*}

\section{Methodology}
\label{method}

This paper explores how representations learnt for object detection change with various privacy-preserving augmentations of the training dataset. We augment the COCO dataset \citep{coco} by face-blurring and face-swapping using an adaptation of StarGAN~\citep{Choi2018stargan}. 
We use these, as well as the raw COCO dataset, to train Faster-RCNN~\citep{ren2016faster}. 
We measure the resulting model's object detection performance on both transformed and original versions. 
Finally, we attempt to measure the bias of the representations of all fine-tuned ResNet50 backbones via the Image Embedding Association Test (iEAT)~\citep{steed2021biases}.

\paragraph{Face Detection and Blurring.}
We augment the COCO dataset so that all faces are blurred and the face colours are scrambled. 
To blur the faces, we take an elliptical area inside each of the bounding boxes returned by the MTCNN face detector~\citep{mtcnn} and perform a Gaussian blur (see Appendix for details).
The intensity of the region of pixels inside each ellipse was shifted by a randomly sampled integer between -80 and +80. This provided a large range of colours, randomising the skin colours of faces in the image, while not saturating the intensity values. 

\paragraph{GAN-based face swapping.}
We develop a pipeline for swapping faces in images using StarGAN~\citep{Choi2018stargan}, detailed in Figure \ref{fig:data_gen}. 
To make a swap, we take a crop of a face detection using MTCNN~\citep{mtcnn} and sample a reference image from UTKFace~\citep{Song2018dataset} with a predefined distribution of gender and/or race\footnote{Definition from UTKFace as in the US census, classified as White, Black, Asian, Indian, and Others}. 
To align the two faces, we first identify facial landmarks with FAN~\citep{bulat2017far}, then generate a matrix transformation to map landmarks to fixed coordinates. 
Using StarGAN, we create an artificial face that has the style of the source face (facial expression) and the texture of the reference face (gender, skin and hair colour). Finally, we warp the generated face to align its landmarks with the face in the source image, remove the background using a pretrained instance segmentation model (Mask-RCNN~\citep{He2017maskcnn}), and paste the segmented face in the original image. 
The detailed procedure is in \ref{app:faceswap}.

\begin{figure*}[t]
	\centering
	\includegraphics[width=\textwidth]{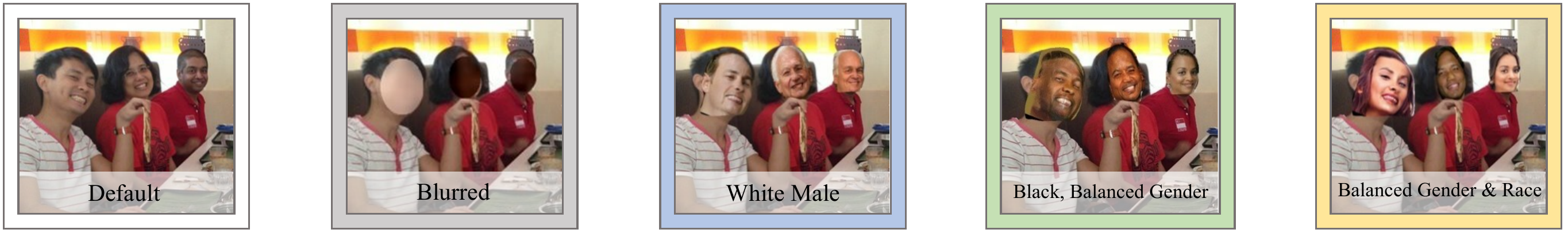}
	\vspace{-1.5em}
   	\caption{\textbf{Example Images.} Five versions of the same COCO image (ID=509864), each taken from one of our datasets. \textit{Default} is the original, \textit{Blurred} was created by blurring detected faces, and \textit{White Male}, \textit{Black \& Balanced Gender}, and \textit{Balanced Gender \& Race} are created using the GAN-based method, augmentations  outlined in Section \ref{method}. \label{fig:datagen}}
\end{figure*}

\paragraph{Bias measure: iEAT test.}
\label{iEAT}
To measure the bias of image representations, we use the iEAT test \citep{steed2021biases}, which is adapted from the social psychology IAT test~\citep{greenwald1998measuring}. 
The test measures the differential association of some target concepts (e.g. ``woman'' vs ``man'') with a set of attributes (e.g. ``maths'' vs ``arts''), in all images (a total of 697). 
We first extract the representations of these visual stimuli using the ResNet backbone in Faster-RCNN models trained on the different datasets. Then, we calculate the cosine distance between the normalized representations of different visual stimuli, and record the $p$-value and effective size $d$ of the null hypothesis significance testing whether the model is biased. 

\paragraph{Privacy-Attribute Leakage: Linear classifier.}
\label{leakage}
We say a model ``leaks'' gender data if it learns different representations for ``man'' and ``woman'' when there has only been label ``person'' in the training set. 
To measure leakage, we train a linear binary classifier on top of the representations of frozen ResNet50 backbones from the Faster-RCNNs. 
We test leakage using two datasets, frame-wise PA-HMDB51~\citep{wang2019privacy} and a Kaggle gender classification dataset~\citep{Orsolini2019dataset}.  
We use this as a proxy for a model's performance on out-of-distribution datasets, and in particular its ability to pick up previously unseen attributes (e.g. the ``blurred'' model has not seen faces).

\section{Results}

\paragraph{Object Detection Performance}
\begin{table}[htb]
\small
\centering
\setlength{\tabcolsep}{3pt}
\caption{\label{Table 1}Object detection performance on COCO and the augmentations of COCO we make. We present mean $AP$, $AP_{75}$ and AP${\Strichmaxerl[1.1]}$ (for mAP of person category) of Faster-RCNN FPN trained (1x schedule) and evaluated on the different combinations datasets. \label{tab:obj_det}}
\begin{tabular}{l l @{\hspace{0.7ex}} c@{\hspace{0.7ex}} c@{\hspace{0.7ex}} c@{\hspace{0.7ex}} @{\hspace{1ex}} c@{\hspace{0.7ex}}
c@{\hspace{0.7ex}} c@{\hspace{0.7ex}}  @{\hspace{1ex}}  c@{\hspace{0.7ex}} c@{\hspace{0.7ex}} c@{\hspace{0.7ex}}  @{\hspace{1ex}}  c @{\hspace{0.7ex}} c@{\hspace{0.7ex}} c@{\hspace{0.7ex}} }
\toprule
    & \multicolumn{1}{r}{\textit{Eval}} & \multicolumn{3}{c}{\textbf{Original}} & \multicolumn{3}{c}{\textbf{Blurred}}  & \multicolumn{3}{c}{\textbf{Black F.}} & \multicolumn{3}{c}{\textbf{Bal. G.\& R.}}\\
    &\textit{Training} & AP & AP$_{75}$ & AP \Strichmaxerl[1.1] &  AP & AP$_{75}$ & AP \Strichmaxerl[1.1] & AP & AP$_{75}$ & AP \Strichmaxerl[1.1] & AP & AP$_{75}$ & AP  \Strichmaxerl[1.1]\\
    \midrule
    \textit{(a)} & Original            & 37.9 & 41.0 & 52.5 & 37.5 & 40.5 & 51.5 & 37.6 & 40.7 & 52.1 & 37.9 & 41.0 & 51.5 \\
    \textit{(b)} & No Faces            & 36.5 & 39.4 & 49.4 & 36.2 & 39.2 & 48.4 & 36.5 & 39.4 & 49.4 & 36.5 & 39.4 & 49.4 \\
    \textit{(c)} & No Persons & 29.7 & 32.8 & 0.0 & 29.6 & 32.7 & 0.0 & 29.7 & 32.7 &  0.0 & 29.7 & 32.7 & 0.0 \\
    \midrule
    \textit{(d)} & Blurred             & 37.8 & 41.1 & 52.5 & 37.6 &\textbf{ 40.9} & \textbf{52.4} & 37.8 & 41.0 & 52.5 & 37.8 & 41.1 & 52.6  \\ 
    \textit{(e)} & White, M            & 37.9 & \textbf{41.2} & 52.6 & 37.7 & \textbf{40.9} & 52.1 & \textbf{37.9} & \textbf{41.2} & 52.6 & 37.9 & \textbf{41.2} & 52.6 \\
    \textit{(f)} & Black, F            & 37.9 & 41.0 & 52.6 & \textbf{37.7} & 40.6 & 51.5 & 37.8 & 40.8 & 52.5 & 37.9 & 41.0 & 52.6 \\
    \textit{(g)} & Black, Bal. G.      & 37.9 & 41.0 & 52.7 & 37.5 & 40.4 & 51.8 & \textbf{37.9 }& 41.0 & 52.8 & 37.9 & 41.0 & \textbf{52.8}    \\
    \textit{(h)} & Bal. G \& R         & \textbf{38.0} & \textbf{41.2} & \textbf{52.8} & 37.6 & 40.8 & 51.7 & \textbf{37.9} & \textbf{41.2} & \textbf{52.8} & \textbf{38.0} & \textbf{41.2} & \textbf{52.8}  \\
\bottomrule
\end{tabular}
\end{table}


The original COCO 2017 dataset contains 118{,}287 training images. 
Removing all images with detected faces yields 86{,}721 training images, while removing all images labelled with the ``person'' class leaves a training set of only 54{,}172 images.
These drops in the number of available training are also echoed in much lower performances in Tab.~\ref{tab:obj_det}(b-c).
Compared to this, even our simple baseline, blurring of faces (row d), works much better and also does not suffer from a loss in performance when evaluating on blurred faces.
Overall, all models trained on modified datasets perform as well as the default model when detecting objects in the COCO dataset. There is no significant drop in AP (<0.1) across any model and measure. 
Each model trained on an modified dataset performs better on that dataset than the default model. In particular, we find the model with balanced gender and race (row h) to perform well across the board. 
This can be attributed to the artifacts around modified faces that the models trained on the modified datasets pick up. 
However, models trained on any modified dataset perform no worse on the default dataset, where no face obfuscation has been performed, than the default model. 
We conclude that given a suitable face obfuscation method, we can train an object detector that does not degrade in performance on real data, despite never having seen a real face.

\paragraph{iEAT Bias Measurements}
The results from the iEAT association test show only minor differences between the default COCO model and our augmentations. The model \textit{Black Female} shows a 20\% decrease in bias against "Arab-Muslim" and \textit{Balanced Gender \& Race} shows a minor increase in bias against "Disability". Biases concerning "(modern) weapons" are statistically significant in \textit{Balanced Gender} and \textit{Balanced Gender \& Race} even though they are not in other models. 
All other differences are either (i) of negligible magnitude (e.g. "Weight"), (ii) are differences of effect size in statistically insignificant results (e.g. in sexuality) or (iii) are differences in the empirically untested intersectional categories (see \ref{app:ieat}). 
This, in combination with the highly significant biases of a randomly initialised RCNN, cast doubt on the validity of the iEAT test.

\paragraph{Privacy-Attribute Leakage Results} 
The accuracy of gender classifiers trained on all models was similar. This suggests that there is no significant reduction in the possibility of decoding gender from the features, even in our backbone \textit{Blurred} which never saw a face during final training. 
It is likely that cues from a person's hair or body must also be altered to successfully obfuscate gender. While obfuscating gender is not a necessary condition for removing bias, we believe further work is required.
Another next step involves pretraining only the weights of ResNet backbones with manipulated input data before finetuning on COCO object detection and repeating these experiments.
\begin{table}[htb]
\footnotesize
    
    \begin{minipage}[t]{.56\linewidth}
      \centering
        \caption{Selected iEAT results (\ref{app:ieat} for full and details).\label{tab:ieat}}
        \vspace{-0.5em}
            \begin{tabular}{l @{\hspace{0.7ex}}l@{\hspace{0.7ex}}l@{\hspace{0.7ex}}l@{\hspace{0.7ex}}l@{\hspace{0.7ex}}l@{\hspace{0.7ex}}l }
                \toprule
                \textit{Model} & Weight & Disability & Arab-Muslim & G.-Science \\
                \midrule
                {Original} & 1.4469$^{***}$  & 1.0590$^{*}$   & 0.7595$^{**}$  & 0.3471$^{*}$  \\ 
                {Blurred}  & 1.4563$^{***}$  & 1.0402$^{*}$   & 0.7484$^{**}$  & 0.3758$^{**}$  \\
                {Bal. G\&R}  & 1.8793$^{***}$ & -0.8023$^{*}$  & 0.5498$^{**}$  & -0.2099$^{*}$  \\
                 \bottomrule
                 \multicolumn{5}{c}{\tiny{*: p<0.1; **: p<0.05; ***: p<0.01}}
               
            \end{tabular}
    \end{minipage} 
    \qquad
    \begin{minipage}[t]{.34\linewidth}
    \setlength{\tabcolsep}{2pt}
      \caption{Selected leakage results.\label{tab:leakage}}
      \vspace{-0.5em}
      \centering
            \begin{tabular}{l c @{\hspace{0.7ex}} c@{\hspace{0.7ex}} c@{\hspace{0.7ex}} }
                \toprule
                \textit{Model} & PA-HMDB51 & M/F \\
                \midrule
                Original  & 80.6 $\pm$ 1.0 & 85.4 $\pm$ 0.8 \\ 
                Blurred  & 80.3 $\pm$ 0.6 & 85.7 $\pm$ 1.5 \\ 
                Bal. G.\&R.  & 80.7 $\pm$ 0.7 & 85.4 $\pm$ 0.7 \\
                \bottomrule
            \end{tabular}
    \end{minipage}%
\end{table}

    

\section{Conclusion}
Does a dataset need real human faces? Not for object detection. In this paper, we show that at least for finetuning on COCO, this is not the case. We find that even when faces are blurred or edited to be more balanced in terms of represented gender and race, the object detector learns just the same. 
We also investigate whether it is possible to de-bias the models using two measures proposed. However, here we report negative and inconsistent findings that at least partially point to potential shortcomings of these measures tested.

\subsection*{Acknowledgement}
The OxAI society is grateful for support from Google Academic Research Credits program.

{\small\bibliographystyle{iclr2021_conference}\bibliography{shortstrings,8_refs}}

\newpage
\newpage 
\appendix
\counterwithin{figure}{section}
\counterwithin{table}{section}
\section{Appendix}

\subsection{Discussion on GDPR}
Under the General Data Protection Regulation (GDPR) biometric data is considered a special category of data (Art. 9.1) and facial images are classified as a type of 'biometric data' (Art 4.14). As such it would be reasonable to conclude that the storage of facial images would likely fall under the GDPR. However, as we have shown, this paper's method is able to remove certain personally identifiable information (PII), leading to what is known as a ‘de-identification’ process~\citep{Hintze2017identify}. By removing PII, whether the data falls under the GDPR and, for example the storage limitation principle (Art. 5.1.e), is brought into question. We believe that as our method reduces PII from images, it also reduces the regulatory requirements for those storing images. This is dependent on if an argument can be made whether the method anonymizes or pseudonymizes the images. If it anonymizes the images, then they would no longer fall under the GDPR (see Recital 26). However, if it pseudonymizes them then it would fall under the GDPR (Art. 4.3b). Regardless of the extent to which our method `de-identifies' subjects, we believe that by reducing PII this provides greater flexibility for researchers to work with computer vision models while adhering to ‘data minimisation’ and ‘storage limitations’ principles where applicable. Therefore, as our method retains information such as facial expression and (to some extent) photorealism it is more favorable than ablation or blurring. We believe that large available datasets could be processed using similar methods before public release to minimise privacy and bias concerns. However, given that under the law it often depends on the specific case in order to provide a legal argument, the fact remains that our method of protecting privacy and preserving performance seems to be of beneficial use.

\subsection{Blurring details}
We apply Gaussian blurring with kernel length $h/6$ (where $h$ is the height of the original bounding box) and a standard deviation of 20 pixels. 

\subsection{Face Swapping with StarGAN}
\label{app:faceswap}
Suppose we want to swap a face in image $I$. We crop out the original face $F$ which can be decomposed as $F = (A, E, L, B)$, where $A$ is appearance and texture, $E$ is the style and facial expressions, $L$ is position of landmarks, and $B$ is background behind the face. Given a predefined gender and/or race setting $p$ for the target dataset, we choose a reference face $R_p = (A_p', E', L', B')$. To align the two faces, we calculate two alignment matrices $\mathcal{M}$, $ \mathcal{M}'$ (calculated such that $\mathcal{M}L=L_0$ and $\mathcal{M}'L'=L_0$, where $L_0$ is a fixed face framework). We apply these to get two aligned face that we denote $F^{(a)}$ and $R^{(a)}$ by $F^{(a)}=\mathcal{M}F,\; R_p^{(a)}=\mathcal{M}'R_p$. Using StarGAN ($\mathcal{T}$), we merge the two faces to create $G^{(a)}_p=\mathcal{T}(F^{(a)}, R_p^{(a)}) = (A'^{(a)}_p, E^{(a)}, L_0, B'^{(a)})$, which retains the style of the original face $E^{(a)}$ but the appearance and background of the reference face $A'^{(a)}_p B'^{(a)}$. The synthesised face is then reverted back to the original position  by using an inverted alignment matrix ($G_p = \mathcal{M}^{-1}G^{(a)}_p=(A_p', E, L, B')$). The background of the artificial face is removed by segmentation (denoted by $\mathcal{S}$) to create $S_p = \mathcal{S}G_p=(A_p', E, L, \emptyset )$. Finally, the face is pasted back to the original image, giving us $I_p$, an altered image with desired gender and skin colour.

\paragraph{Hyper-parameters of StarGAN:}
Most hyper-parameters are default values from the pretrained models: MTCNN, FAN, StarGAN, and Mask-RCNN. \\
We only change the margin parameter of StarGAN to $0.8$ and the threshold of MTCNN to $0.5$.

\subsubsection{Gender Classification}
\paragraph{Male/Female Kaggle dataset} We train the frozen ResNet50 backbones for 20 epochs, using Adam optimizer \citep{kingma2017adam} with learning rate $0.001$ and a batch size of 16, and use 3-fold cross validation.

\paragraph{PA-HMDB51 dataset} PA-HMDB51 contains videos labelled by actions (such as "brush\_hair") and frame-by-frame labels of privacy attributes including skin colour and gender. We extract frames containing only one "male"/"female" gender label to create a dataset of images labelled by gender and use this to train a gender classifier.
As with the Male/Female Kaggle dataset, we train the frozen ResNet50 backbones for 1 epoch, using Adam optimizer \citep{kingma2017adam} with learning rate 0.001 and a batch size of 16, and use 3-fold cross-validation.

\subsection{Leakage results}
\label{sec:leakage-app}
\begin{table}[!h]
\footnotesize
      \caption{Full leakage results.}
      \label{tab:leakage-app}
      \centering
            \begin{tabular}{l c @{\hspace{0.7ex}} c@{\hspace{0.7ex}} c@{\hspace{0.7ex}} }
                \toprule
                \textit{Model} & PA-HMDB51 & M/F \\
                \midrule
                Original  & 80.6 $\pm$ 1.0 & 85.4 $\pm$ 0.8 \\
                Blurred  & 80.3 $\pm$ 0.6 & 85.7 $\pm$ 1.5 \\ 
                White M  & 80.2 $\pm$ 0.7 & 85.7 $\pm$ 0.6 \\
                Black F  & 79.9 $\pm$ 0.6 & 85.8 $\pm$ 0.3 \\
                Bal. G.\&R.  & 80.7 $\pm$ 0.7 & 85.4 $\pm$ 0.7 \\
                \bottomrule
            \end{tabular}
\end{table}

\subsection{iEAT Test Results}
\label{app:ieat}
The iEAT test measures bias along a number of variables, but we focus on the ones that we identify as \textit{social} biases, and only discuss statistically significant results. There is a decrease in the \textit{Gender-Science}, \textit{Arab-Muslim} and \textit{Weight} categories when comparing \textbf{Balanced-intersectional} to \textbf{Default}.Comparing \textbf{Black-female} to \textbf{Default}, we see a 10\% to 20\% decrease of effective size in \textit{Gender-Science}, \textit{Arab-Muslim} and \textit{Disability}. \textbf{White-male} increases the bias in all categories with significant results, \textit{Weight} and \textit{Disability}, compared to \textbf{Default}. Similarly, \textbf{Blurred} increases \textit{Gender-Science} and \textit{Weight} biases, while slightly decreasing along the \textit{Disability} and \textit{Arab-Muslim} categories.\\
For completeness, we include results for a randomly initialised RCNN, as well as an RCNN backbone pretrained on ImageNet. While the results of the randomly initialised one can not be interpreted, the ImageNet-pretrained one shows the smallest bias in the \textit{Weight} and \textit{Arab-Muslim} categories, and the highest in \textit{Disability}. Models trained on altered datasets show similar bias results, suggesting that data preprocessing has a small effect. Changing the layer from which we take representations extraction layer shows no difference in the results\\
An important thing to note is that as mentioned in the original iEAT paper~\citep{steed2021biases}, the IAT bias test on humans, where this test is adapted from, did not include the bias tests on intersectionality. Therefore, the results for intersectional bias have not been empirically compared and supported. The fact that even the randomly initialised model has significant biases in some categories means that the intersectional biases are likely false positives. \\
In addition, the tests use a small number of visual stimuli (normally less than 10 for each attribute and concept) and sample 3000-10000 partitions from the union of 2 attributes and 2 concepts, so statistical bias could result from the small selection of images.

\midsepremove
\begin{table}
\scriptsize
\setlength{\tabcolsep}{3pt}

\newcolumntype{a}{>{\columncolor[gray]{0.8}}l}

\caption{iEAT bias test full results: p value and effective size D.
Abbreviations: \textbf{Rnd}: Randomly initialised, \textbf{IDf}: Imagenet pretrained, \textbf{Df}: default COCO pretrained, \textbf{Bl}: Blurred, \textbf{BF}: Black female, \textbf{WM}: White male, \textbf{Bal-R}: balanced along race, \textbf{Bal-G}: balanced along gender, \textbf{Bal-I}: balanced intersectionally.}

\begin{tabular}{l | llallllll}
                             & \multicolumn{9}{c}{\textbf{p-value / Effect-size}}              \\                                                                                                     \\
\toprule
                                                            & \textbf{Rnd}                   & \textbf{IDf}                  & \textbf{Df}                   & \textbf{Bl}                                          & \textbf{BF}                    & \textbf{WM}                   & \textbf{Bal-R}               & \textbf{Bal-G}                    & \textbf{Bal-I}     \\

\midrule
\multirow{2}{*}{\textbf{Insect-Flower}                      }& {\color[HTML]{807F7F} 1.000}  & {\color[HTML]{807F7F} 0.988}  & {\color[HTML]{807F7F} 1.000}  & {\color[HTML]{807F7F} 1.000}  & {\color[HTML]{807F7F} 1.000}   & {\color[HTML]{807F7F} 1.000}  & {\color[HTML]{807F7F} 1.000}    & {\color[HTML]{807F7F} 1.000}  & {\color[HTML]{807F7F} 1.000}  \\
                                                             & {\color[HTML]{807F7F} -0.945} & {\color[HTML]{807F7F} -0.553} & {\color[HTML]{807F7F} -0.970} & {\color[HTML]{807F7F} -1.125}                        & {\color[HTML]{807F7F} -0.963}  & {\color[HTML]{807F7F} -1.111} & {\color[HTML]{807F7F}   -1.086} & {\color[HTML]{807F7F} -1.033} & {\color[HTML]{807F7F} -1.030} \\ \midrule
\multirow{2}{*}{\textbf{Gender-Science}                     }& {\color[HTML]{807F7F} 0.378}  & {\color[HTML]{807F7F} 0.219}  & 0.060*                        & 0.046**                                              & 0.084*                         & {\color[HTML]{807F7F} 0.119}  & {\color[HTML]{807F7F} 0.133   } & 0.051*                        & 0.065*                        \\
                                                             & {\color[HTML]{807F7F} 0.069}  & {\color[HTML]{807F7F} 0.177}  & 0.347                         & 0.376                                                & 0.313                          & {\color[HTML]{807F7F} 0.267}  & {\color[HTML]{807F7F} 0.250}    & 0.369                         & 0.337                         \\ \midrule
\multirow{2}{*}{\textbf{Gender-Career}                      }& {\color[HTML]{807F7F} 0.133}  & 0.081*                        & {\color[HTML]{807F7F} 0.735}  & {\color[HTML]{807F7F} 0.779}                         & {\color[HTML]{807F7F} 0.746}   & {\color[HTML]{807F7F} 0.848}  & {\color[HTML]{807F7F} 0.781}  & {\color[HTML]{807F7F} 0.666}  & {\color[HTML]{807F7F} 0.798}  \\
                                                             & {\color[HTML]{807F7F} 0.253}  & 0.305                         & {\color[HTML]{807F7F} -0.140} & {\color[HTML]{807F7F} -0.174}                        & {\color[HTML]{807F7F} -0.154}  & {\color[HTML]{807F7F} -0.234} & {\color[HTML]{807F7F} -0.172}    & {\color[HTML]{807F7F} -0.102} & {\color[HTML]{807F7F} -0.196} \\ \midrule
\multirow{2}{*}{\textbf{Disability}                         }& {\color[HTML]{807F7F} 0.786}  & 0.014**                       & 0.071*                        & 0.071*                                               & 0.071*                         & 0.071*                        & 0.071*                         & 0.071*                        & 0.071*                        \\
                                                             & {\color[HTML]{807F7F} -0.567} & 1.285                         & 1.059                         & 1.040                                                & 0.974                          & 1.107                         &                            1.101 & 1.056                         & 1.129                         \\ \midrule
\multirow{2}{*}{\textbf{Asian}                              }& {\color[HTML]{807F7F} 0.118}  & {\color[HTML]{807F7F} 0.233}  & {\color[HTML]{807F7F} 0.325}  & {\color[HTML]{807F7F} 0.310}                         & {\color[HTML]{807F7F} 0.022**} & {\color[HTML]{807F7F} 0.529}  & {\color[HTML]{807F7F} 0.172}     & {\color[HTML]{807F7F} 0.563}  & {\color[HTML]{807F7F} 0.452}  \\
                                                             & {\color[HTML]{807F7F} 0.740}  & {\color[HTML]{807F7F} 0.483}  & {\color[HTML]{807F7F} 0.294}  & {\color[HTML]{807F7F} 0.335}                         & {\color[HTML]{807F7F} 1.147}   & {\color[HTML]{807F7F} -0.061} & {\color[HTML]{807F7F} 0.593}   & {\color[HTML]{807F7F} -0.093} & {\color[HTML]{807F7F} 0.080}  \\ \midrule
\multirow{2}{*}{\textbf{Arab-Muslim}                        }& {\color[HTML]{807F7F} 0.158}  & 0.089*                        & 0.048**                       & 0.044**                                              & 0.084*                         & {\color[HTML]{807F7F} 0.100}  & {\color[HTML]{807F7F} 0.108}   & 0.050**                       & 0.049**                       \\
                                                             & {\color[HTML]{807F7F} 0.450}  & 0.620                         & 0.760                         & 0.748                                                & 0.628                          & {\color[HTML]{807F7F} 0.595}  & {\color[HTML]{807F7F} 0.567}   & 0.746                         & 0.754                         \\ \midrule
\multirow{2}{*}{\textbf{Age}                                }& {\color[HTML]{807F7F} 0.963}  & {\color[HTML]{807F7F} 0.343}  & {\color[HTML]{807F7F} 0.513}  & {0.543}                         & {\color[HTML]{807F7F} 0.530}   & {\color[HTML]{807F7F} 0.538}  & {\color[HTML]{807F7F} 0.472}    & {\color[HTML]{807F7F} 0.502}  & {\color[HTML]{807F7F} 0.447}  \\
                                                             & {\color[HTML]{807F7F} -1.000} & {\color[HTML]{807F7F} 0.277}  & {\color[HTML]{807F7F} -0.022} & {\color[HTML]{807F7F} -0.096}                        & {\color[HTML]{807F7F} -0.057}  & {\color[HTML]{807F7F} -0.064} & {\color[HTML]{807F7F} 0.033}    & {\color[HTML]{807F7F} -0.003} & {\color[HTML]{807F7F} 0.087}  \\ \midrule
\multirow{2}{*}{\textbf{Weight}                             }& \textless{}1e-3***            & 0.003***                      & \textless{}1e-3***            & \textless{}1e-3***                                   & \textless{}1e-3***             & \textless{}1e-3***            &                           <1e-3*** & \textless{}1e-3***            & \textless{}1e-3***            \\
                                                             & 1.828                         & 1.117                         & 1.447                         & 1.456                                                & 1.493                          & 1.523                         &                           1.358 & 1.484                         & 1.410                         \\ \midrule
\multirow{2}{*}{\textbf{Weapon (Modern)}                    }& {\color[HTML]{807F7F} 0.168}  & {\color[HTML]{807F7F} 0.181}  & {\color[HTML]{807F7F} 0.384}  & {\color[HTML]{807F7F} 0.723}                         & {\color[HTML]{807F7F} 0.127}   & {\color[HTML]{807F7F} 0.795}  & {\color[HTML]{807F7F} 0.489}    & 0.037**                       & 0.035**                       \\
                                                             & {\color[HTML]{807F7F} 0.568}  & {\color[HTML]{807F7F} 0.579}  & {\color[HTML]{807F7F} 0.180}  & {\color[HTML]{807F7F} -0.374}                        & {\color[HTML]{807F7F} 0.672}   & {\color[HTML]{807F7F} -0.515} & {\color[HTML]{807F7F} 0.035 }   & 1.026                         & 1.010                         \\ \midrule
\multirow{2}{*}{\textbf{Weapon}                             }& {\color[HTML]{807F7F} 0.421}  & {\color[HTML]{807F7F} 0.981}  & {\color[HTML]{807F7F} 0.120}  & {\color[HTML]{807F7F} 0.215}                         & {\color[HTML]{807F7F} 0.374}   & {\color[HTML]{807F7F} 0.740}  & {\color[HTML]{807F7F} 0.427}    & {\color[HTML]{807F7F} 0.457}  & {\color[HTML]{807F7F} 0.707}  \\
                                                             & {\color[HTML]{807F7F} 0.118}  & {\color[HTML]{807F7F} -1.138} & {\color[HTML]{807F7F} 0.700}  & {\color[HTML]{807F7F} 0.477}                         & {\color[HTML]{807F7F} 0.192}   & {\color[HTML]{807F7F} -0.418} & {\color[HTML]{807F7F} 0.110}   & {\color[HTML]{807F7F} 0.062}  & {\color[HTML]{807F7F} -0.347} \\ \midrule
\multirow{2}{*}{\textbf{Skin-Tone}                          }& \textless{}1e-3***            & {\color[HTML]{807F7F} 0.446}  & {\color[HTML]{807F7F} 0.355}  & {\color[HTML]{807F7F} 0.341}                         & {\color[HTML]{807F7F} 0.395}   & {\color[HTML]{807F7F} 0.302}  & {\color[HTML]{807F7F} 0.339}    & {\color[HTML]{807F7F} 0.385}  & {\color[HTML]{807F7F} 0.377}  \\
                                                             & 1.531                         & {\color[HTML]{807F7F} 0.075}  & {\color[HTML]{807F7F} 0.151}  & {\color[HTML]{807F7F} 0.108}                         & {\color[HTML]{807F7F} 0.087}   & {\color[HTML]{807F7F} 0.163}  & {\color[HTML]{807F7F} 0.198}     & {\color[HTML]{807F7F} 0.076}  & {\color[HTML]{807F7F} 0.101}  \\ \midrule
\multirow{2}{*}{\textbf{Sexuality}                          }& {\color[HTML]{807F7F} 0.371}  & {\color[HTML]{807F7F} 0.694}  & {\color[HTML]{807F7F} 0.606}  & {\color[HTML]{807F7F} 0.558}                         & {\color[HTML]{807F7F} 0.565}   & {\color[HTML]{807F7F} 0.583}  & {\color[HTML]{807F7F} 0.532}    & {\color[HTML]{807F7F} 0.628}  & {\color[HTML]{807F7F} 0.644}  \\
                                                             & {\color[HTML]{807F7F} 0.163}  & {\color[HTML]{807F7F} -0.244} & {\color[HTML]{807F7F} -0.137} & {\color[HTML]{807F7F} -0.084}                        & {\color[HTML]{807F7F} -0.075}  & {\color[HTML]{807F7F} -0.100} & {\color[HTML]{807F7F} -0.039}    & {\color[HTML]{807F7F} -0.147} & {\color[HTML]{807F7F} -0.178} \\ \midrule
\multirow{2}{*}{\textbf{Religion}                           }& {\color[HTML]{807F7F} 0.217}  & {\color[HTML]{807F7F} 0.338}  & {\color[HTML]{807F7F} 0.206}  & {\color[HTML]{807F7F} 0.276}                         & {\color[HTML]{807F7F} 0.269}   & {\color[HTML]{807F7F} 0.222}  & {\color[HTML]{807F7F} -0.292}    & {\color[HTML]{807F7F} 0.227}  & {\color[HTML]{807F7F} 0.259}  \\
                                                             & {\color[HTML]{807F7F} 0.437}  & {\color[HTML]{807F7F} 0.229}  & {\color[HTML]{807F7F} 0.455}  & {\color[HTML]{807F7F} 0.337}                         & {\color[HTML]{807F7F} 0.344}   & {\color[HTML]{807F7F} 0.429}  & {\color[HTML]{807F7F} 0.322}    & {\color[HTML]{807F7F} 0.412}  & {\color[HTML]{807F7F} 0.358}  \\ \midrule
\multirow{2}{*}{\textbf{Race}                               }& 0.021**                       & {\color[HTML]{807F7F} 0.933}  & {\color[HTML]{807F7F} 0.648}  & {\color[HTML]{807F7F} 0.635}                         & {\color[HTML]{807F7F} 0.642}   & {\color[HTML]{807F7F} 0.669}  & {\color[HTML]{807F7F} 0.705}    & {\color[HTML]{807F7F} 0.680}  & {\color[HTML]{807F7F} 0.649}  \\
                                                             & 1.137                         & {\color[HTML]{807F7F} -0.856} & {\color[HTML]{807F7F} -0.268} & {\color[HTML]{807F7F} -0.306}                        & {\color[HTML]{807F7F} -0.252}  & {\color[HTML]{807F7F} -0.324} & {\color[HTML]{807F7F} -0.384}    & {\color[HTML]{807F7F} -0.352} & {\color[HTML]{807F7F} -0.271} \\ \midrule
\multirow{2}{*}{\textbf{Native}                             }& {\color[HTML]{807F7F} 0.489}  & {\color[HTML]{807F7F} 0.998}  & {\color[HTML]{807F7F} 0.821}  & {\color[HTML]{807F7F} 0.634}                         & {\color[HTML]{807F7F} 0.862}   & {\color[HTML]{807F7F} 0.793}  & {\color[HTML]{807F7F} 0.706 }    & {\color[HTML]{807F7F} 0.885}  & {\color[HTML]{807F7F} 0.430}  \\
                                                             & {\color[HTML]{807F7F} 0.021}  & {\color[HTML]{807F7F} -1.323} & {\color[HTML]{807F7F} -0.474} & {\color[HTML]{807F7F} -0.174}                        & {\color[HTML]{807F7F} -0.568}  & {\color[HTML]{807F7F} -0.419} & {\color[HTML]{807F7F} -0.270}   & {\color[HTML]{807F7F} -0.613} & {\color[HTML]{807F7F} 0.094}  \\ 
\midrule                                                  
\midrule                                                    
\multirow{2}{*}{\textbf{Intersectional-Valence-WMWF}        }& 0.032**                       & 0.045**                       & 0.017**                       & 0.011**                                              & 0.040**                        & 0.038**                       &   0.041** & 0.014**                       & 0.012**                       \\
                                                             & 0.585                         & 0.533                         & 0.661                         & 0.708                                                & 0.554                          & 0.561                         &   0.557 & 0.683                         & 0.711                         \\ \midrule
\multirow{2}{*}{\textbf{Intersectional-Valence-WMBM}        }& \textless{}1e-3***            & {\color[HTML]{807F7F} 0.581}  & {\color[HTML]{807F7F} 0.257}  & {\color[HTML]{807F7F} 0.375}                         & {\color[HTML]{807F7F} 0.380}   & {\color[HTML]{807F7F} 0.280}  & {\color[HTML]{807F7F} 0.315 }& {\color[HTML]{807F7F} 0.312}  & {\color[HTML]{807F7F} 0.134}  \\
                                                             & 1.052                         & {\color[HTML]{807F7F} -0.072} & {\color[HTML]{807F7F} 0.215}  & {\color[HTML]{807F7F} 0.103}                         & {\color[HTML]{807F7F} 0.095}   & {\color[HTML]{807F7F} 0.188}  & {\color[HTML]{807F7F} 0.163 }   & {\color[HTML]{807F7F} 0.163}  & {\color[HTML]{807F7F} 0.354}  \\ \midrule
\multirow{2}{*}{\textbf{Intersectional-Valence-WMBF}        }& \textless{}1e-3***            & {\color[HTML]{807F7F} 0.313}  & {\color[HTML]{807F7F} 0.139}  & {\color[HTML]{807F7F} 0.237}                         & {\color[HTML]{807F7F} 0.156}   & {\color[HTML]{807F7F} 0.133}  & {\color[HTML]{807F7F} 0.172 }& {\color[HTML]{807F7F} 0.191}  & {\color[HTML]{807F7F} 0.109}  \\
                                                             & 1.163                         & {\color[HTML]{807F7F} 0.156}  & {\color[HTML]{807F7F} 0.352}  & {\color[HTML]{807F7F} 0.234}                         & {\color[HTML]{807F7F} 0.323}   & {\color[HTML]{807F7F} 0.354}  & {\color[HTML]{807F7F} 0.303}    & {\color[HTML]{807F7F} 0.288}  & {\color[HTML]{807F7F} 0.391}  \\ \midrule
\multirow{2}{*}{\textbf{Intersectional-Valence-WFBM}        }& \textless{}1e-3***            & 0.044**                       & 0.007***                      & 0.019**                                              & 0.038**                        & 0.023**                       &                     0.026** & 0.012**                       & 0.002***                      \\
                                                             & 1.313                         & 0.536                         & 0.736                         & 0.648                                                & 0.555                          & 0.621                         &                         0.604 & 0.704                         & 0.927                         \\ \midrule
\multirow{2}{*}{\textbf{Intersectional-Valence-WFBF}        }& \textless{}1e-3***            & 0.007***                      & \textless{}1e-3***            & 0.002***                                             & 0.004***                       & 0.003***                      &                    0.005*** & \textless{}1e-3***            & \textless{}1e-3***            \\
                                                             & 1.361                         & 0.765                         & 0.987                         & 0.878                                                & 0.836                          & 0.828                         &                         0.817 & 0.927                         & 1.080                         \\ \midrule
\multirow{2}{*}{\textbf{Intersectional-Valence-FM}          }& {\color[HTML]{807F7F} 0.459}  & {\color[HTML]{807F7F} 0.256}  & {\color[HTML]{807F7F} 0.134}  & {\color[HTML]{807F7F} 0.128}                         & {\color[HTML]{807F7F} 0.263}   & {\color[HTML]{807F7F} 0.220}  & {\color[HTML]{807F7F} 0.208}    & {\color[HTML]{807F7F} 0.129}  & 0.071*                        \\
                                                             & {\color[HTML]{807F7F} 0.022}  & {\color[HTML]{807F7F} 0.150}  & {\color[HTML]{807F7F} 0.252}  & {\color[HTML]{807F7F} 0.252}                         & {\color[HTML]{807F7F} 0.146}   & {\color[HTML]{807F7F} 0.173}  & {\color[HTML]{807F7F} 0.187}    & {\color[HTML]{807F7F} 0.259}  & 0.327                         \\ \midrule
\multirow{2}{*}{\textbf{Intersectional-Valence-BW}          }& \textless{}1e-3***            & 0.080*                        & 0.006***                      & 0.021**                                              & 0.024**                        & 0.013**                       &       0.018** & 0.011**                       & 0.001***                      \\
                                                             & 1.224                         & 0.319                         & 0.558                         & 0.457                                                & 0.442                          & 0.494                         &             0.462 & 0.511                         & 0.678                         \\ \midrule
\multirow{2}{*}{\textbf{Intersectional-Valence-BFBM}        }& {\color[HTML]{807F7F} 0.845}  & {\color[HTML]{807F7F} 0.787}  & {\color[HTML]{807F7F} 0.582}  & {\color[HTML]{807F7F} 0.603}                         & {\color[HTML]{807F7F} 0.712}   & {\color[HTML]{807F7F} 0.628}  & {\color[HTML]{807F7F} 0.606}    & {\color[HTML]{807F7F} 0.583}  & {\color[HTML]{807F7F} 0.460}  \\
                                                             & {\color[HTML]{807F7F} -0.338} & {\color[HTML]{807F7F} -0.257} & {\color[HTML]{807F7F} -0.068} & {\color[HTML]{807F7F} -0.086}                        & {\color[HTML]{807F7F} -0.181}  & {\color[HTML]{807F7F} -0.113} & {\color[HTML]{807F7F} -0.091}    & {\color[HTML]{807F7F} -0.072} & {\color[HTML]{807F7F} 0.030}  \\ \midrule
\multirow{2}{*}{\textbf{Intersectional-Gender-Science-WMWF} }& {\color[HTML]{807F7F} 0.980}  & {\color[HTML]{807F7F} 0.957}  & {\color[HTML]{807F7F} 0.850}  & {\color[HTML]{807F7F} 0.548}                         & {\color[HTML]{807F7F} 0.781}   & {\color[HTML]{807F7F} 0.817}  & {\color[HTML]{807F7F}  0.828}   & {\color[HTML]{807F7F} 0.874}  & {\color[HTML]{807F7F} 0.889}  \\
                                                             & {\color[HTML]{807F7F} -0.649} & {\color[HTML]{807F7F} -0.539} & {\color[HTML]{807F7F} -0.333} & {\color[HTML]{807F7F} -0.041}                        & {\color[HTML]{807F7F} -0.243}  & {\color[HTML]{807F7F} -0.295} & {\color[HTML]{807F7F} -0.298 }  & {\color[HTML]{807F7F} -0.372} & {\color[HTML]{807F7F} -0.396} \\ \midrule
\multirow{2}{*}{\textbf{Intersectional-Gender-Science-WMBM} }& 0.006***                      & {\color[HTML]{807F7F} 0.424}  & {\color[HTML]{807F7F} 0.135}  & {\color[HTML]{807F7F} 0.474}                         & {\color[HTML]{807F7F} 0.626}   & {\color[HTML]{807F7F} 0.521}  & {\color[HTML]{807F7F} 0.320}   & {\color[HTML]{807F7F} 0.221}  & {\color[HTML]{807F7F} 0.116}  \\
                                                             & 0.768                         & {\color[HTML]{807F7F} 0.058}  & {\color[HTML]{807F7F} 0.353}  & {\color[HTML]{807F7F} 0.018}                         & {\color[HTML]{807F7F} -0.109}  & {\color[HTML]{807F7F} -0.018} & {\color[HTML]{807F7F} 0.155}    & {\color[HTML]{807F7F} 0.249}  & {\color[HTML]{807F7F} 0.377}  \\ \midrule
\multirow{2}{*}{\textbf{Intersectional-Gender-Science-WMBF} }& \textless{}1e-3***      &     {\color[HTML]{807F7F} 0.125}  & {0.042**}                        & {\color[HTML]{807F7F} 0.225}                         & {\color[HTML]{807F7F} 0.141}   & {\color[HTML]{807F7F} 0.101}  & 0.042** & 0.098*                        & 0.032**                       \\
                                                             & 1.017                         &          {\color[HTML]{807F7F} 0.365}  & 0.548                         & {\color[HTML]{807F7F} 0.253}                         & {\color[HTML]{807F7F} 0.357}   & {\color[HTML]{807F7F} 0.418}  &   0.543 & 0.410                         & 0.577                         \\ \midrule
\multirow{2}{*}{\textbf{Intersectional-Gender-Science-MF}   }& {\color[HTML]{807F7F} 0.742}  & {\color[HTML]{807F7F} 0.685}  & {\color[HTML]{807F7F} 0.594}  & {\color[HTML]{807F7F} 0.316}                         & {\color[HTML]{807F7F} 0.327}   & {\color[HTML]{807F7F} 0.354}  & {\color[HTML]{807F7F} 0.436}    & {\color[HTML]{807F7F} 0.649}  & {\color[HTML]{807F7F} 0.603}  \\
                                                             & {\color[HTML]{807F7F} -0.155} & {\color[HTML]{807F7F} -0.108} & {\color[HTML]{807F7F} -0.057} & {\color[HTML]{807F7F} 0.108}                         & {\color[HTML]{807F7F} 0.104}   & {\color[HTML]{807F7F} 0.081}  & {\color[HTML]{807F7F} 0.038}   & {\color[HTML]{807F7F} -0.084} & {\color[HTML]{807F7F} -0.066} \\ \midrule
\multirow{2}{*}{\textbf{Intersectional-Gender-Career-WMWF}  }& {\color[HTML]{807F7F} 0.950}  & {\color[HTML]{807F7F} 0.688}  & {\color[HTML]{807F7F} 0.926}  & {\color[HTML]{807F7F} 0.862}                         & {\color[HTML]{807F7F} 0.833}   & {\color[HTML]{807F7F} 0.871}  & {\color[HTML]{807F7F} 0.874}    & {\color[HTML]{807F7F} 0.933}  & {\color[HTML]{807F7F} 0.939}  \\
                                                             & {\color[HTML]{807F7F} -0.524} & {\color[HTML]{807F7F} -0.162} & {\color[HTML]{807F7F} -0.463} & {\color[HTML]{807F7F} -0.352}                        & {\color[HTML]{807F7F} -0.315}  & {\color[HTML]{807F7F} -0.366} & {\color[HTML]{807F7F} -0.368}   & {\color[HTML]{807F7F} -0.474} & {\color[HTML]{807F7F} -0.487} \\ \midrule
\multirow{2}{*}{\textbf{Intersectional-Gender-Career-WMBM}  }& {\color[HTML]{807F7F} 0.877}  & {\color[HTML]{807F7F} 0.275}  & 0.076*                        & 0.045**                                              & {\color[HTML]{807F7F} 0.122}   & 0.045**                       & {\color[HTML]{807F7F} 0.119 }   & {\color[HTML]{807F7F} 0.207}  & {\color[HTML]{807F7F} 0.145}  \\
                                                             & {\color[HTML]{807F7F} -0.372} & {\color[HTML]{807F7F} 0.194}  & 0.456                         & 0.539                                                & {\color[HTML]{807F7F} 0.375}   & 0.528                         & {\color[HTML]{807F7F} 0.383}   & {\color[HTML]{807F7F} 0.269}  & {\color[HTML]{807F7F} 0.342}  \\ \midrule
\multirow{2}{*}{\textbf{Intersectional-Gender-Career-WMBF}  }& 0.019**                       & {\color[HTML]{807F7F} 0.874}  & {\color[HTML]{807F7F} 0.667}  & {\color[HTML]{807F7F} 0.704}                         & {\color[HTML]{807F7F} 0.526}   & {\color[HTML]{807F7F} 0.735}  & {\color[HTML]{807F7F} 0.628}    & {\color[HTML]{807F7F} 0.498}  & {\color[HTML]{807F7F} 0.521}  \\
                                                             & 0.659                         & {\color[HTML]{807F7F} -0.365} & {\color[HTML]{807F7F} -0.141} & {\color[HTML]{807F7F} -0.172}                        & {\color[HTML]{807F7F} -0.021}  & {\color[HTML]{807F7F} -0.204} & {\color[HTML]{807F7F} 0.628}    & {\color[HTML]{807F7F} 0.008}  & {\color[HTML]{807F7F} -0.009} \\ \midrule
\multirow{2}{*}{\textbf{Intersectional-Gender-Career-MF}    }& {\color[HTML]{807F7F} 0.734}  & {\color[HTML]{807F7F} 0.743}  & {\color[HTML]{807F7F} 0.618}  & {\color[HTML]{807F7F} 0.476}                         & {\color[HTML]{807F7F} 0.463}   & {\color[HTML]{807F7F} 0.521}  & {\color[HTML]{807F7F} 0.577}    & {\color[HTML]{807F7F} 0.675}  & {\color[HTML]{807F7F} 0.638}  \\
                                                             & {\color[HTML]{807F7F} -0.147} & {\color[HTML]{807F7F} -0.145} & {\color[HTML]{807F7F} -0.069} & {\color[HTML]{807F7F} 0.010}                         & {\color[HTML]{807F7F} 0.020}   & {\color[HTML]{807F7F} -0.008} & {\color[HTML]{807F7F} -0.043}   & {\color[HTML]{807F7F} -0.102} & {\color[HTML]{807F7F} -0.080} \\ 
   
\bottomrule
\end{tabular}
\end{table}

\newpage
\subsection{Example-Image Gallery}
\begin{figure*}[h]
	\centering
	\includegraphics[width=\textwidth]{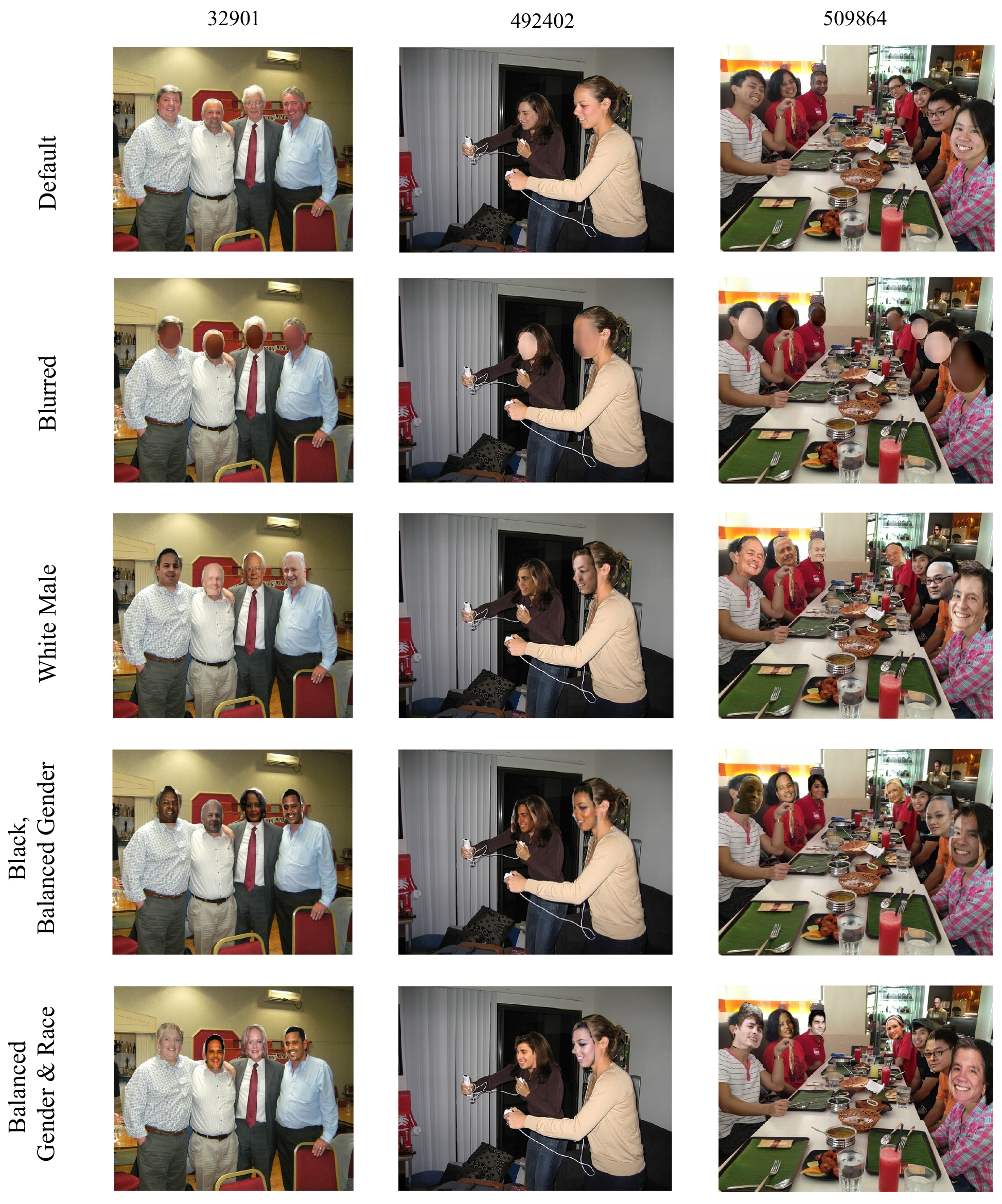}
	\caption{A selection of example images used for training. They are labelled by their COCO IDs, such as "$32901$", as well as the dataset they belong to. Details about the creation of each dataset can be found in \ref{method}.
	\label{fig:augmented_data}}
\end{figure*}
\end{document}